\documentclass{article}
\usepackage{acra}

\usepackage[sorting=none, style=nature, backend=biber]{biblatex}
\title{Textile Anomaly Detection: Evaluation of the State-of-the-Art for Automated Quality Inspection of Carpet}
\author{
    Briony Forsberg$^{1*}$, Henry Williams$^{1}$, Bruce MacDonald$^{1}$, \\\textbf{Tracy Chen$^{2}$, Kirstine Hulse$^{2}$}\\
    Centre for Automation and Robotic Engineering Science, The University of Auckland, New Zealand$^{1}$.\\
    Bremworth Ltd, Auckland, New Zealand$^{2}$.\\
  \texttt{bfor519@aucklanduni.ac.nz}$^*$ \\
}

\begin{document}

\maketitle

\begin{abstract}
In this study, state-of-the-art unsupervised detection models were evaluated for the purpose of automated anomaly inspection of wool carpets. A custom dataset of four unique types of carpet textures was created to thoroughly test the models and their robustness in detecting subtle anomalies in complex textures. Due to the requirements of an inline inspection system in a manufacturing use case, the metrics of importance in this study were accuracy in detecting anomalous areas, the number of false detections, and the inference times of each model for real-time performance. Of the evaluated models, the student-teacher network based methods were found on average to yield the highest detection accuracy and lowest false detection rates. When trained on a multi-class dataset the models were found to yield comparable if not better results than single-class training. Finally, in terms of detection speed, with exception to the generative model, all other evaluated models were found to have comparable inference times on a GPU, with an average of 0.16s per image. On a CPU, most of these models typically produced results between 1.5 to 2 times the respective GPU inference times. 
\end{abstract}

\section{Introduction}

Manual quality inspections and tests are still commonly conducted to check for irregularities and impurities in products throughout textile manufacturing processes \cite{WONG201847}. These tests can be time consuming, laborious and suffer from operator fatigue and inattentiveness, which leads to unreliable and inaccurate results. Additionally, these manual assessments are subjective and results can differ from person to person.  When proposing their own online monitoring system for woven fabrics in \cite{anagnostopoulos2001computer}, the authors showed that, in some cases, only about 70\% of the defects could be detected by the most highly trained inspectors. 
\par
The aim of the work presented in this paper was to examine the accuracy and robustness of existing state-of-the-art automated anomaly detection methods on a custom wool carpet dataset. Existing methods for anomaly detection are often developed for generic industrial inspection purposes \cite{30_peng21,37_nagy22,52_bionda22,55_roth22,58_rudolph21,67_fastflow,68_reverse_dist,69_stfpm,71_padim} and tested on the common publicly available datasets \cite{mvtec1,mvtec2,steel1,steel2,tilda}. However, before being implemented in a real-world setting, these existing methods need to be tailored and tuned on specific data to assess their ability in yielding the required detection accuracy and acceptable false alarm rate. This paper will thus explore which methods show promise for an automated carpet inspection system, for future work in developing such a system. \par
Challenges that exist with visual inspection of defects include diverse sizes, varying environmental conditions, and complex features of the products being inspected. Depending on the type of fibre used (i.e. synthetics, cotton, wool), textile defects can vary in shape, size, and colour and can have ambiguous edges and low contrast. The textile products themselves can be patterned and range from fine to coarse texture. To this end, the models and algorithms that are chosen for this specific application must meet the following criteria: 
\begin{itemize}
    \item Robust: to detect objects of all shapes and sizes on products of varying features  
    \item Fast: to aid real-time, in-line quality assessment 
    \item Scalable: to adapt to new, unseen products and anomalies 
\end{itemize}
\par
Due to these challenges and the complex diversity of both fabric textures and defects, the authors of \cite{34_liu19} discuss how many defect detection methods are still sub-optimal. 

\section{Background}

The following sections aim to provide an overview of anomaly detection in the textile manufacturing industry, including the requirements of real-time monitoring systems and associated challenges. Furthermore, this section describes how defects can occur in wool tufted carpets and what these can look like. 

\subsection{Visual Anomaly Detection in Textiles}

In their review, the authors of \cite{28_rasheed20} note that due to the limited number of publicly accessible fabric and textile datasets, it is difficult for researchers to find an optimal anomaly detection method for their specific application. Because studies are often conducted with different databases, different parameters and varied imaging systems, the validity and reliability of methods is far from objective. \par
Additionally, the specific industrial use cases affect the evaluation of which methods are best \cite{WONG201847}. Firstly, the inference time requirement will be influenced by the maximum speed of the product at the point of inspection. Another consideration is the width of the product and how many image patches will be required from discretisation before resolution affects detection accuracy. Finally, depending on the environmental factors, for example lighting and vibration, the expectation of model robustness will vary from project to project. \par
Textile products can be visually assessed based on colour mismatch, unintentional texture variation, and pattern inconsistency. According to \cite{29_pereira22}, several challenges are inherent to the inspection of textiles including intentional variability, textural volume and shadows, and dark shades of colour. Vision systems used to analyse the quality of textiles must first “learn” the product, the properties of the yarn, the colours and tolerable imperfections. This often requires “teaching”, through labelling many images, and “learning”, through supervised learning techniques. \par

\subsection{Woollen Yarn and Carpet Manufacturing}

\subsubsection{Wool yarns}

According to \cite{27_pereira18}, the quality of the final textile product is directly related to the quality of the yarn. The appearance and durability of the fabric is affected, as well as the productivity and efficiency of the manufacturing functions \cite{woolmark}.\par
Visual anomalies of wool yarns include, but are not limited to, contamination and debris, count variation (thin and thick areas), twist variation, streaky yarn (colour and/or texture inconsistency), hairiness, and neps (tangled fibres) \cite{woolmark,wronz}. 
All of these issues at the yarn stage will go on to affect the appearance of the final product \cite{woolmark}. \par

\subsubsection{Tufted carpet}

The main types of tufted carpet are loop pile and cut pile \cite{carpet_89}. By adjusting machine parameters at tufting, to control stitch rate and pile height for example, and specifying particular properties for the yarns, these two types of carpets can then expand into a whole range of unique and complex textures and patterns. \par
As discussed in the previous section, defects in carpets can arise due to the yarns used or from the manufacturing of the carpet (tufting). Visual anomalies in carpet can include high/low lines, texture variation, and colour or pattern inconsistencies \cite{carpet_93}. See Figure 1 for examples of defects that can occur in tufted carpet.\par

\begin{figure*}[hbt!]
\centering
\includegraphics[width=1.0\textwidth]{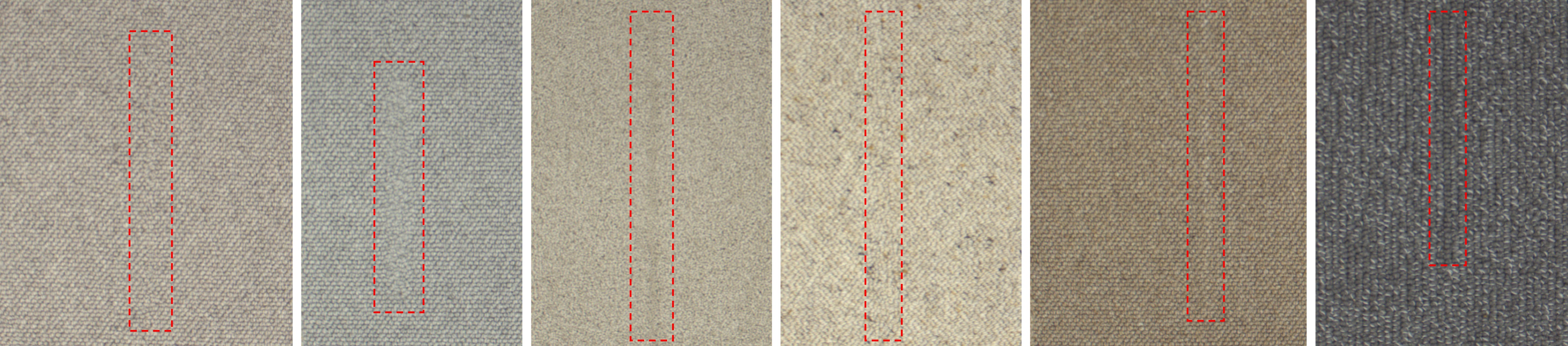}
\caption{Examples of carpet defects.}
\label{fig:Real_Defects}
\end{figure*}

Much of the literature in anomaly detection for textiles specifically focuses on woven cotton or synthetic fabrics \cite{30_peng21, 31_jun21, 33_zhan22, 34_liu19, 35_hu20, 56_ccelik14, 79_li13}. However, unlike these "flat" fabrics, viewing angle and lighting conditions can significantly vary the appearance of anomalies in the three dimensional nature of the tufted carpet pile. This makes automatic visual inspection of carpets a complex task. In addition, the automated inspection system needs to understand which irregularities are intentional and which are not, to avoid false detections. \par
Many fabric defect detection algorithms that have been tested on carpets refer to the use of the MVTec Textures dataset \cite{30_peng21,84_ren22}, whereby there is a single instance of a carpet texture, one of six classes that make up the Textures dataset \cite{mvtec1, mvtec2}. Another popular textiles dataset is the TILDA Texture Database \cite{tilda} which was originally used to develop methods to recognise and distinguish varying textures. This dataset has a total of eight texture classes and seven defect classes for each. \par
Neither the single MVTec carpet class nor the generic woven fabric textures of the TILDA dataset are very representative of tufted wool carpets. In developing an effective detection system for wool carpets, these existing methods must be trained and evaluated on a specific, custom dataset before an approach can be selected and fine-tuned. \par

\section{Anomaly Detection Methods}

A range of methods exist for anomaly detection. These can be broken down into representation-based methods, supervised model-based methods, and unsupervised model-based methods. Each of these approaches are discussed in the following sections.

\subsection{Representation-Based Methods}

Prior to deep learning methods, features within images were extracted using representation-based methods for highlighting whether anomalies were present or not. Based on the type of algorithm used, these methods can be classified into three main groups: statistical approaches, model-based approaches, and spectral approaches \cite{72_jiang18}. \par
When compared against deep learning methods, representation-based methods are always much faster (inference within at least half the time \cite{31_jun21}) but yield lower detection rates and higher false alarm rates. In evaluating their deep learning-based fabric defect detection method, the authors of \cite{31_jun21} tested five representation-based methods, that used feature extraction such as 
Gabor filters or Fourier and wavelet transforms. In comparison to the deep learning methods, the methods that used Gabor filters, which were most popular for detecting defects in fabrics, were found to be sensitive to small texture variations and exhibited high false detection rates. \par

\subsection{Supervised Learning Methods}

Popular supervised learning-based object detection algorithms include Faster R-CNN \cite{fast_rcnn}, YOLO \cite{yolo}, or SSD \cite{ssd}. However, when choosing the basis for their detection network in \cite{34_liu19}, the authors note that networks such as these would likely yield poor detection accuracy because fabric defects vary widely in scale and aspect ratio. Additionally, the authors of \cite{38_tao18} note that while these data-driven, deep learning methods achieve high accuracy and real-time detection they require large, annotated datasets. However, in real-world manufacturing applications, it is very difficult to create a large, comprehensive dataset that includes all possible variations of both products and defects.\par
The limitation of requiring a lot of manually labelled data to learn how to detect anomalies is the major disadvantage of supervised models that aim to learn what “bad” looks like. Semi-supervised and unsupervised models however do not require so much effort in data labelling as the training data would be defect-free. To this end, these models would learn what “good” looks like and, when presented with images containing defects, recognise when something looks incorrect.  In the manufacturing industry, it is typically much easier to obtain images of defect-free products than defective images. 

\subsection{Unsupervised Learning Methods}

Unsupervised methods for anomaly detection can be broadly grouped into three main approaches: generative methods, pretrained and feature embedding methods, and student-teacher methods. The next few sections will briefly explain each of these approaches as well as recent literature that employed these methods for anomaly detection and their respective advantages and disadvantages. 

\subsubsection{Generative methods}

Generative models, when used for anomaly detection, aim to recreate input images sans anomalous areas. The founding idea being that anomalies cannot be generated by the trained networks because they do not exist in the training datasets. Generative adversarial networks (GANs) are one major instance of this type of model \cite{39_goodfellow20}. \par
A deep convolutional GAN (DCGAN) method for detecting defects in fabrics was proposed in \cite{35_hu20}. In addition to the typical Discriminator (D) and Generator (G) networks of a GAN, this method introduced an Inverter component (CNN model) that enabled the complete model to recreate a defective image without the anomalous areas. By subtracting this reconstructed image from the original, a residual map could be created that highlights defective regions.\par
In similar work, \cite{70_ganomaly} also employed an encoder network (like the Inverter in \cite{35_hu20}) to translate a true image to the latent space before being used as the input for the Generator. The main distinguishing point between the two approaches being how each individual model was trained. In \cite{35_hu20}, the Encoder was optimised independently from the other networks. \par
The disadvantage of GANs is that, because both the Discriminator and Generator are trained separately to minimise the loss but are adversarial by nature, it is easy for GANs to suffer mode collapse, whereby D and G prevent each other from learning. \par
In response to this, the authors of \cite{57_wei21} chose to use Autoencoder (AE) and Variational Autoencoder (VAE) models for the development of a real-time fabric defect detection system. Unlike GANs, the Encoder and Decoder networks of AE models are trained simultaneously. Here, the encoder attempts to reduce an input image to a strongly compressed, encoded form and the Decoder then attempts to recreate the image. For their AE-based methods, in selecting an appropriate loss function to assess reconstruction quality, the authors of \cite{57_wei21, 52_bionda22} used variations of the Structural Similarity (SSIM) index which the authors considered to be more robust at describing textures than other popular loss functions. \par
Overall, the effectiveness of generative models is largely dependent on the latent space that the extracted features are mapped too and this space is very hard to optimise. Also, in general, low-level feature reconstruction from high-level representation is a difficult task \cite{68_reverse_dist}. With images of high variability and complex patterns, even close reconstructions can still yield large errors, resulting in noisy anomaly maps. This is the main reason for why GAN-based defect detection methods discretise the original images into many smaller patches, to simplify the structure to be recreated. 

\subsubsection{Pretrained and Feature Embedding Methods}

These methods utilise the learned weights of popular deep learning models that have been trained on large-scale, labelled datasets for image classification tasks. For the purpose of anomaly detection, these pretrained models are used as feature extractors, whereby feature vectors from select intermediate layers of the models are concatenated to form discriminative embeddings. Unlike generative methods, these methods can have no training loss, and instead just fine-tune the parameters characterising the embedding (latent) space from the training data. The methods that do have training losses for optimisation employ another model, in addition to the pretrained model, in order to produce more discriminative embeddings \cite{59_bergmann20}.\par
The method in \cite{58_rudolph21} utilises normalising flow (NF) networks to transform the extracted features from the pretrained network to a latent space and then calculate a likelihood. Due to the bijective mapping nature of NFs, these likelihoods can then be backpropagated to optimise the NF network parameters. The training objective is to minimise the likelihoods for anomaly-free image patches such that they are as close as possible to zero. \cite{67_fastflow} also utilises NF networks to learn transformations between data distributions. During inference, the features of anomalous images should be "out of distribution" and hence have higher likelihoods, which are the anomaly scores per pixel. \par 
While structured around a student-teacher learning paradigm (discussed in the next section), \cite{59_bergmann20} treats it as a regression problem.  An encoding network is trained to produce accurate discriminative embeddings (or "descriptors") by a total loss that includes a hard negative mining loss (ensuring similar image patches have closer embeddings than very different patches) and a correlation loss (to reduce correlation and redundancy between embeddings). A decoding network is then trained to reduce the regression errors and prediction variance between their feature embeddings and that of the encoder.\par
The method in \cite{71_padim} aimed to preserve information from different semantic levels and resolutions by concatenating the embedding vectors, extracted from the feature layers of the pretrained CNN (Wide ResNet-50 pretrained on ImageNet), for each patch within an image. These concatenated embeddings were then mapped to a Gaussian distribution and the distribution parameters for each image patch were learnt and fine-tuned over the entire training dataset.\par

\subsubsection{Student-teacher methods}
    
Student-Teacher methods can be seen as a combination of generative methods and pretrained methods. The concept of the Student-Teacher approach is to utilise the knowledge of a deep learning network (the "teacher") that has been pretrained on a large and comprehensive dataset while training another network (the "student") on a domain specific dataset. The underlying idea being that the complex and multi-scale features that the Teacher network has been fine-tuned to extract helps the Student to learn multi-scale representations from its specific training data. Typically the Student networks share the same architecture as the Teacher networks to avoid information loss.\par 
In \cite{69_stfpm} feature vectors from specific layers in the Teacher network (ResNet pretrained on ImageNet) and the Student network were selected for comparison in determining the overall loss. The loss function used was the average of the l2-distance losses at each patch position within the image, and then a weighted average across the different feature scales. In \cite{68_reverse_dist}, instead of raw images, the Student takes in the Teacher's one-class embedding output as input and aims to reproduce the multi-scale anomaly maps from the intermediate layers. The purpose of the one-class bottleneck embedding module between the Teacher and Student networks was to reduce the multi-scale features to an extremely low dimensional space such that anomalous data is abandoned.\par
The authors of \cite{30_peng21} proposed an approach where normal, defect-free images were input into a pretrained Teacher network (VGG19) to extract features of what “good” looks like. An AE-based network then learned to reconstruct the feature vector that was extracted from the Teacher network. Finally, the Student network, a simple CNN, is trained to mimic the behaviour of the Teacher network.\par
The authors of \cite{59_bergmann20} trained an ensemble of Student networks on anomaly-free data, so that the final likelihood map is the average from across the ensemble and yields higher confidence. Another difference with the method in \cite{59_bergmann20} is that the pretrained network (ResNet-18 pretrained on ImageNet) was used to teach the Teacher network. By decoding the Teacher embeddings back to image patch features and comparing these with those produced by the pretrained network, the authors found this produced more appropriate discriminative embeddings.\par

\section{Methodology}

We aim to evaluate current state-of-the-art anomaly detection methods on carpet defects in order to determine which type of method works best for the specific use case. The following sections describe how these existing techniques were bench-marked to compare their behavior in this domain.

\subsection{Dataset}

\begin{figure}[h]
\centering
\includegraphics[width=0.45\textwidth]{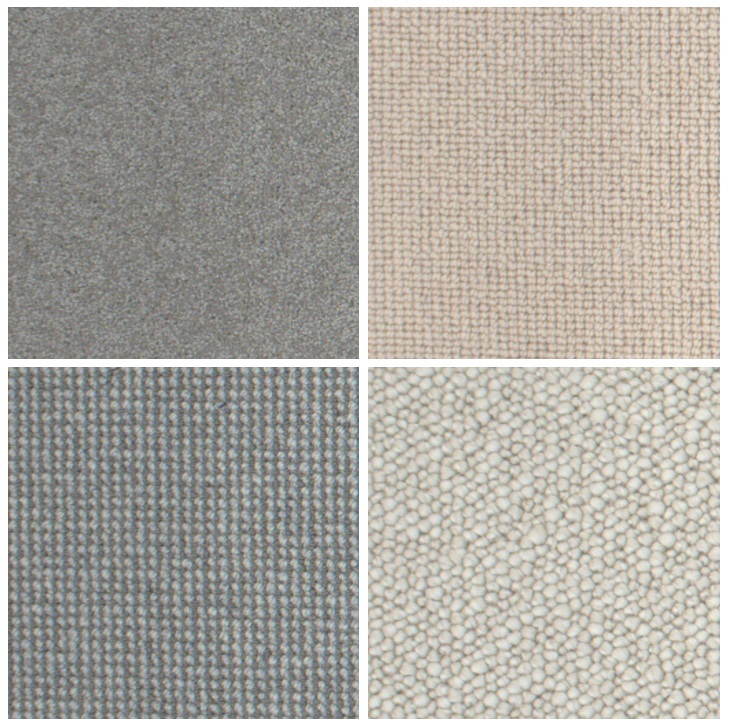}
\caption{Four types of distinct carpet textures, clockwise from top-left: cut pile (C1), loop pile regular structure (C2), loop pile irregular structure (C4), loop pile multi-colour regular structure (C3)}
\label{fig:Custom_Dataset}
\end{figure}

To thoroughly evaluate the detection methods specifically for carpets, four image classes were chosen to encompass the wide variety of textures and patterns that are possible. Examples of each of these classes can be seen in Figure \ref{fig:Custom_Dataset}. Carpet 1 (C1) is a single colour cut pile which, while uniform in pattern, the texture appears variable due to the shading that is caused by the twist of the piles. Carpet 2 (C2) is a single colour, regular-patterned loop pile while Carpet 3 (C3) is multi-coloured. Finally, Carpet 4 (C4) is a multi-colour, irregularly patterned loop pile. All four carpets are 100\% wool, thereby the fibre type is constant. \par
The images were collected with a Basler area scan camera (acAacA4096-30uc) that was mounted 1200mm above carpet samples that contributed to the four carpet classes. These images were then cropped and flipped around the horizontal and vertical to create the full dataset. With this camera setup, images of the carpet classes from various points of view were captured, and not just from straight above. This should allow for the models to learn the acceptable distortions of the textures when viewed from different angles. \par
The training datasets for each of the four carpet structures consists of 300 anomaly-free images of 512 x 512 pixels. For testing and validation, there are 80 images per class which contain different types of anomalies. For the purpose of this work, to yield an adequate number of samples for images, these anomalies were created by hand to replicate as close as possible the true defects that can appear (see Figure 1). \par
The database also contains the segmentation masks for all the images with defects. These masks are completely black except for areas of white pixels which indicate the defective areas, and the rest of the pixels are black. These images are not used during the training process and are solely for evaluating the models. \par

\subsection{State-of-the-Art Approaches}

The aim for this evaluation study was to use existing state-of-the-art unsupervised anomaly detection methods and assess which work best on carpet textures. However, the results of several of the methods discussed in the previous section, that were specifically developed for fabric anomaly detection, could not be reproduced \cite{30_peng21, 31_jun21, 34_liu19, 35_hu20}. Instead, methods that had provided code online were selected with at least one method being from each of the three unsupervised learning categories:
\begin{itemize}
    \item \textbf{Generative:} AE with CW-SSIM \cite{52_bionda22}
    \item \textbf{Feature Embedding:} FastFlow \cite{67_fastflow}, Patch Distribution Modelling (PaDiM) \cite{71_padim}, PatchCore \cite{55_roth22}
    \item \textbf{Student-Teacher:} Reverse Distillation \cite{68_reverse_dist}, Student Teacher Feature Pyramid Matching (STFPM) \cite{69_stfpm}
\end{itemize}

Some of these selected models could be found in the recently established deep learning library, Anomalib \cite{anomalib}. Anomalib is comprised of state-of-the-art models specifically for image based anomaly detection. One purpose of Anomalib was to enable easy benchmarking of the detection algorithms, with many of the ready-to-use implementations having been tested on the MVTec dataset \cite{mvtec1, mvtec2} to allow for comparison. \par
All the parameters of the evaluated models were derived from their respective origin literature. With exception, for comparison sake, of the backbone networks used for both PatchCore \cite{55_roth22} and Reverse Distillation \cite{68_reverse_dist}. In both of these methods, WideResNet-50 was found to yield the best accuracy during evaluation while the smaller ResNet-18 architecture was used by the other methods. The authors of \cite{68_reverse_dist, 55_roth22} acknowledged however that while WideResNet-50 produced the best results, it came at the expense of inference time and the results from ResNet-18 were comparable and were produced in much less time. \par

\subsection{Evaluation Metrics}

According to \cite{31_jun21}, the performance of defect detection algorithms are generally evaluated by three criteria: (1) the detection rate, represents the sensitivity of the detection model, (2) the false alarm rate, reflects the robustness of the model, (3) model efficiency, denotes the feasibility of the model for industrial application. \par
Some fundamental performance metrics for segmentation models have been defined in Table \ref{table:metrics}. For this study, two overall metrics were used: F1-Score and the Area Under the Per Region Overlap (AUPRO).

\renewcommand{\arraystretch}{1.25}
\begin{table}[h]
\footnotesize 
\caption{(a) Confusion matrix definitions, (b) Evaluation metrics for classification and segmentation models}
\label{table:metrics}
\resizebox{\columnwidth}{!}{%
\begin{tabular}{llcl}
\multicolumn{4}{c}{(a)} \\ \cline{3-4} 
\multicolumn{2}{l|}{} & \multicolumn{1}{c|}{\textbf{Predicted anomaly}} & \multicolumn{1}{c|}{\textbf{Predicted normal}} \\ \hline
\multicolumn{2}{|l|}{\textbf{\begin{tabular}[c]{@{}l@{}}Ground truth\\ anomaly\end{tabular}}} & \multicolumn{1}{c|}{\begin{tabular}[c]{@{}c@{}}Correct anomaly\\ (TP: True Positive)\end{tabular}} & \multicolumn{1}{c|}{\begin{tabular}[c]{@{}c@{}}Missed anomaly\\ (FN: False Negative)\end{tabular}} \\ \hline
\multicolumn{2}{|l|}{\textbf{\begin{tabular}[c]{@{}l@{}}Ground truth\\ normal\end{tabular}}} & \multicolumn{1}{c|}{\begin{tabular}[c]{@{}c@{}}Incorrect anomaly\\ (FP: False Postive)\end{tabular}} & \multicolumn{1}{c|}{\begin{tabular}[c]{@{}c@{}}Correct normal\\ (TN: True Negative)\end{tabular}} \\ \hline
\multicolumn{4}{c}{(b)} \\ \hline
(1) & \multicolumn{2}{c}{Recall = $\frac{TP}{TP + FN}$} & \begin{tabular}[c]{@{}l@{}}Ratio of all anomalous \\ pixels correctly classified\end{tabular} \\
(2) & \multicolumn{2}{c}{Precision = $\frac{TP}{TP + FP}$} & \begin{tabular}[c]{@{}l@{}}Ratio of all classified \\ anomalous pixels that are\\ correct\end{tabular} \\
(3) & \multicolumn{2}{c}{FPR = $\frac{FP}{FP + TN}$} & \begin{tabular}[c]{@{}l@{}}Ratio of all normal pixels\\ incorrectly classified as\\ anomalous\end{tabular} \\
(4) & \multicolumn{2}{c}{F1 = $\frac{2 . Recall . Precision}{Recall + Precision}$} & \begin{tabular}[c]{@{}l@{}}Harmonic mean between\\ Recall and Precision\end{tabular} \\ \hline
\end{tabular}%
}
\end{table}

\begin{figure*}[h]
\centering
\includegraphics[width=1.0\textwidth]{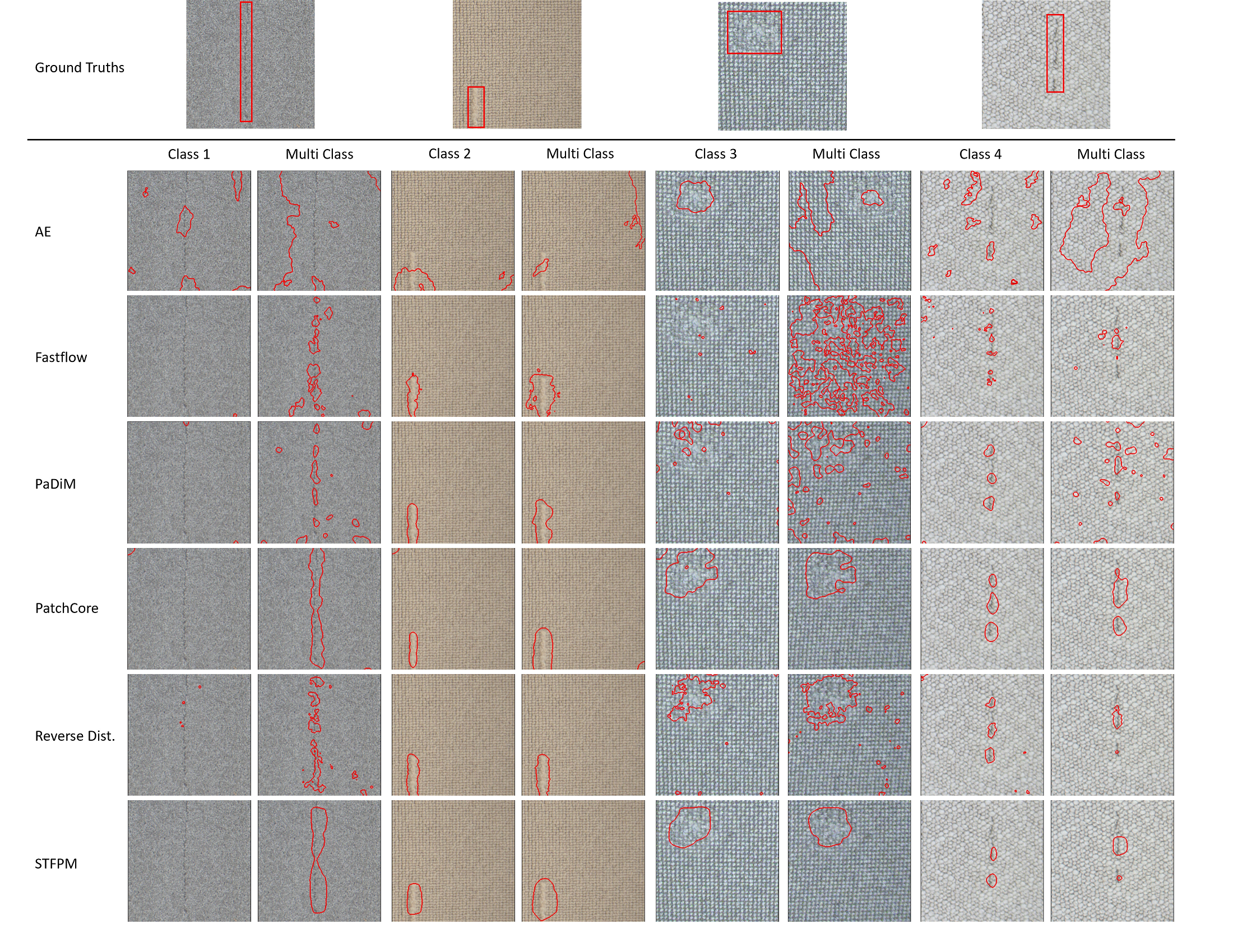}
\caption{Results of each detection method for each carpet class. Comparison between the results from each model trained on a single carpet class versus a combined dataset (multi class) is also shown.}
\label{fig:Detection_Results}
\end{figure*}

The F1-Score, also known as the Dice Coefficient, is a popular segmentation metric. The F1-Score is the harmonic mean between both Recall and Precision, such that a high score indicates good prediction quantity and quality. AUPRO is a threshold independent metric and is used for per-pixel measurement. AUPRO treats every anomalous region equally regardless of size. It consists of plotting, for each connected component, a curve for the mean values of the correctly classified pixels as a function of the false positive rate between 0 and 0.3 \cite{71_padim}. The final score is the normalised integral of this curve. A high PRO score means that both large and small anomalies were well localised by a model. \par

\section{Results and Discussion}



\renewcommand{\arraystretch}{1.25}
\begin{table*}[ht]
\scriptsize 
\centering
\caption{Pixel-level detection results for each method trained on (a) single-class datasets and (b) a multi-class dataset}
\label{table:results}
\begin{tabular}{lcccccccc}

\multicolumn{9}{c}{(a)} \\ \hline
\multicolumn{1}{|l||}{} & \multicolumn{4}{c||}{AUPRO} & \multicolumn{4}{c|}{F1-Score} \\ \cline{2-9} 
\multicolumn{1}{|l||}{\multirow{-2}{*}{Model}} & \multicolumn{1}{c|}{C1} & \multicolumn{1}{c|}{C2} & \multicolumn{1}{c|}{C3} & \multicolumn{1}{c||}{C4} & \multicolumn{1}{c|}{C1} & \multicolumn{1}{c|}{C2} & \multicolumn{1}{c|}{C3} & \multicolumn{1}{c|}{C4} \\ \hhline{-||----||----}

\multicolumn{1}{|l||}{AE} & \multicolumn{1}{c|}{0.486} & \multicolumn{1}{c|}{0.720} & \multicolumn{1}{c|}{0.321} & \multicolumn{1}{c||}{0.644} & \multicolumn{1}{c|}{\cellcolor[HTML]{C0C0C0}} & \multicolumn{1}{c|}{\cellcolor[HTML]{C0C0C0}} & \multicolumn{1}{c|}{\cellcolor[HTML]{C0C0C0}} & \multicolumn{1}{c|}{\cellcolor[HTML]{C0C0C0}} \\ \hhline{-||----||----}

\multicolumn{1}{|l||}{FastFlow} & \multicolumn{1}{c|}{0.854} & \multicolumn{1}{c|}{0.975} & \multicolumn{1}{c|}{0.730} & \multicolumn{1}{c||}{0.878} & \multicolumn{1}{c|}{0.662} & \multicolumn{1}{c|}{0.691} & \multicolumn{1}{c|}{0.503} & \multicolumn{1}{c|}{0.603} \\ \hhline{-||----||----}

\multicolumn{1}{|l||}{PaDiM} & \multicolumn{1}{c|}{\textbf{0.920}} & \multicolumn{1}{c|}{0.973} & \multicolumn{1}{c|}{0.803} & \multicolumn{1}{c||}{0.930} & \multicolumn{1}{c|}{0.568} & \multicolumn{1}{c|}{0.650} & \multicolumn{1}{c|}{0.518} & \multicolumn{1}{c|}{0.576} \\ \hhline{-||----||----}

\multicolumn{1}{|l||}{PatchCore} & \multicolumn{1}{c|}{0.840} & \multicolumn{1}{c|}{0.970} & \multicolumn{1}{c|}{0.760} & \multicolumn{1}{c||}{0.900} & \multicolumn{1}{c|}{0.592} & \multicolumn{1}{c|}{0.641} & \multicolumn{1}{c|}{0.546} & \multicolumn{1}{c|}{0.571} \\ \hhline{-||----||----}

\multicolumn{1}{|l||}{Reverse Dist.} & \multicolumn{1}{c|}{0.832} & \multicolumn{1}{c|}{0.972} & \multicolumn{1}{c|}{\textbf{0.860}} & \multicolumn{1}{c||}{0.950} & \multicolumn{1}{c|}{0.555} & \multicolumn{1}{c|}{0.692} & \multicolumn{1}{c|}{0.609} & \multicolumn{1}{c|}{0.646} \\ \hhline{-||----||----}

\multicolumn{1}{|l||}{STFPM} & \multicolumn{1}{c|}{0.732} & \multicolumn{1}{c|}{\textbf{0.980}} & \multicolumn{1}{c|}{\textbf{0.860}} & \multicolumn{1}{c||}{\textbf{0.960}} & \multicolumn{1}{c|}{\textbf{0.679}} & \multicolumn{1}{c|}{\textbf{0.719}} & \multicolumn{1}{c|}{\textbf{0.646}} & \multicolumn{1}{c|}{\textbf{0.707}} \\ \hhline{-||----||----}

\multicolumn{9}{c}{(b)} \\ \hline
\multicolumn{1}{|l||}{} & \multicolumn{4}{c||}{AUPRO} & \multicolumn{4}{c|}{F1-Score} \\ \cline{2-9} 
\multicolumn{1}{|l||}{\multirow{-2}{*}{Model}} & \multicolumn{1}{c|}{C1} & \multicolumn{1}{c|}{C2} & \multicolumn{1}{c|}{C3} & \multicolumn{1}{c||}{C4} & \multicolumn{1}{c|}{C1} & \multicolumn{1}{c|}{C2} & \multicolumn{1}{c|}{C3} & \multicolumn{1}{c|}{C4} \\ \hhline{-||----||----}

\multicolumn{1}{|l||}{AE} & \multicolumn{1}{c|}{0.243} & \multicolumn{1}{c|}{0.468} & \multicolumn{1}{c|}{0.186} & \multicolumn{1}{c||}{0.424} & \multicolumn{1}{c|}{\cellcolor[HTML]{C0C0C0}} & \multicolumn{1}{c|}{\cellcolor[HTML]{C0C0C0}} & \multicolumn{1}{c|}{\cellcolor[HTML]{C0C0C0}} & \multicolumn{1}{c|}{\cellcolor[HTML]{C0C0C0}} \\ \hhline{-||----||----}

\multicolumn{1}{|l||}{FastFlow} & \multicolumn{1}{c|}{0.792} & \multicolumn{1}{c|}{0.956} & \multicolumn{1}{c|}{0.485} & \multicolumn{1}{c||}{0.840} & \multicolumn{1}{c|}{0.407} & \multicolumn{1}{c|}{0.656} & \multicolumn{1}{c|}{0.180} & \multicolumn{1}{c|}{0.373} \\ \hhline{-||----||----}

\multicolumn{1}{|l||}{PaDiM} & \multicolumn{1}{c|}{0.725} & \multicolumn{1}{c|}{0.956} & \multicolumn{1}{c|}{0.520} & \multicolumn{1}{c||}{0.808} & \multicolumn{1}{c|}{0.250} & \multicolumn{1}{c|}{0.672} & \multicolumn{1}{c|}{0.223} & \multicolumn{1}{c|}{0.393} \\ \hhline{-||----||----}

\multicolumn{1}{|l||}{PatchCore} & \multicolumn{1}{c|}{0.761} & \multicolumn{1}{c|}{0.963} & \multicolumn{1}{c|}{0.767} & \multicolumn{1}{c||}{0.962} & \multicolumn{1}{c|}{0.310} & \multicolumn{1}{c|}{0.655} & \multicolumn{1}{c|}{0.588} & \multicolumn{1}{c|}{0.712} \\ \hhline{-||----||----}

\multicolumn{1}{|l||}{Reverse Dist.} & \multicolumn{1}{c|}{\textbf{0.857}} & \multicolumn{1}{c|}{0.972} & \multicolumn{1}{c|}{\textbf{0.895}} & \multicolumn{1}{c||}{\textbf{0.964}} & \multicolumn{1}{c|}{\textbf{0.485}} & \multicolumn{1}{c|}{0.704} & \multicolumn{1}{c|}{\textbf{0.649}} & \multicolumn{1}{c|}{\textbf{0.755}} \\ \hhline{-||----||----}

\multicolumn{1}{|l||}{STFPM} & \multicolumn{1}{c|}{0.672} & \multicolumn{1}{c|}{\textbf{0.974}} & \multicolumn{1}{c|}{0.790} & \multicolumn{1}{c||}{0.963} & \multicolumn{1}{c|}{0.296} & \multicolumn{1}{c|}{\textbf{0.767}} & \multicolumn{1}{c|}{0.536} & \multicolumn{1}{c|}{0.752} \\ \hhline{-||----||----}

\end{tabular}
\end{table*}


\renewcommand{\arraystretch}{1.25}
\begin{table}[]
\scriptsize 
\caption{Results for the Reverse Distillation and STFPM methods trained on the multi-class dataset of image sizes 256x256 and 512x512 pixels}
\label{table:image_size}
\resizebox{\columnwidth}{!}{%
\begin{tabular}{|l|l||cc||cc|}
\hline
\multirow{2}{*}{Model} & \multirow{2}{*}{\begin{tabular}[c]{@{}l@{}}Metric \\ Level\end{tabular}} & \multicolumn{2}{c||}{AUPRO} & \multicolumn{2}{c|}{F1-Score} \\ \hhline{~~||--||--} 
 &  & \multicolumn{1}{c|}{256} & 512 & \multicolumn{1}{c|}{256} & 512 \\ \hhline{--||--||--} 
\multirow{2}{*}{\begin{tabular}[c]{@{}l@{}}Reverse \\ Dist.\end{tabular}} & Image & \multicolumn{1}{c|}{-} & - & \multicolumn{1}{c|}{0.912} & \textbf{0.932} \\ \hhline{~-||--||--} 
 & Pixel & \multicolumn{1}{c|}{0.713} & \textbf{0.863} & \multicolumn{1}{c|}{\textbf{0.623}} & 0.601 \\ \hhline{--||--||--} 
\multirow{2}{*}{STFPM} & Image & \multicolumn{1}{c|}{-} & - & \multicolumn{1}{c|}{0.907} & \textbf{0.942} \\ \hhline{~-||--||--} 
 & Pixel & \multicolumn{1}{c|}{0.547} & \textbf{0.812} & \multicolumn{1}{c|}{0.478} & \textbf{0.566} \\ \hhline{--||--||--} 
\end{tabular}%
}
\end{table}

From the results in Table \ref{table:results}a, the student-teacher networks typically produced the higher accuracy scores than the other models. This could be due to the fact that the student-teacher models do not overwrite the learnt information from the pretrained networks. This pretrained knowledge is retained while another network learns specific features from the training dataset. As can be seen in Figure 3, the FastFlow and PaDiM models typically produced noisier segmentation maps than the student-teacher models. In practice, this would lead to inaccurate localisation of the anomalies and a higher number of false alarms. As seen in Table \ref{table:results}a, these two models were typically less accurate than STFPM by 10\%. \par
As for the generative AE-based model, this was the worst performing model for this dataset. Due to the complexity and scale of the carpet textures, the reconstructed images were typically slightly "off" from the input images. As discussed in \cite{68_reverse_dist}, small deviations in the reconstructed images result in noise anomaly maps. These noisy maps meant that thresholding to yield low FPR and high TPR was difficult and to retain flagged anomalous areas, a high number of false alarms would also be present. \par
With respect to the classes, as can be seen in Figure 3, the models typically produced uniformly good results for Class 2, which was a single colour, regularly patterned carpet. This is in contrast with the results for Class 3, the multi colour, regularly patterned carpet, the results for which were more varied. \par
Interestingly, even though the pattern of Class 1 is uniform, due to the twist of the cut pile, the texture appears to be variable due to areas of shading. This can be seen in Figure 3, where poor detection of a linear anomaly in Class 1 occurred when the models were only trained on Class 1 data. \par
To investigate this further, Table \ref{table:results}b documents the performance of each model when trained on a multi-class dataset, which combined all four classes of carpet in one. It can be seen in Figure 3 that, when trained on multi class data, the linear anomaly is largely detected by all models except for the AE. This could be because the apparent texture of Class 1 is random and the textures of the other three classes are more structured and predictable. Thus the models were able to learn from the structured classes, for example linear features and local inconsistencies, and transfer this knowledge for more accurate detection in Class 1. \par
Overall, the multi-class-trained models produced comparable results to the single-class models, showing promise for having a single trained model that can be used for anomaly detection on multiple carpets instead of loading parameters specifically for each carpet. \par
Another area of evaluation was to assess how model accuracy is affected by input image patch size. As discussed in the next section, the larger the image patch size, the fewer the number of images to be processed at once, resulting in a proportionally lower inference time required for detection. \par
All of the evaluated models resize the input images to 256x256 pixels. As the size of the image patches increases, the amount of information lost to this downscaling process also increases. However, as can be seen in Table \ref{table:image_size}, for the two best-performing models, larger image patches of 512x512 pixels result in higher detection accuracies than image patches of 256x256 pixels. This could be due to two reasons. Firstly, at the scale of four pixels to one, the resizing process removes noise from the images but not vital anomaly information. Secondly, the larger image patches that have been cropped from the full camera image have a larger field of view and thus a more global perspective for understanding normal and anomalous structures. \par

\subsection{Real-Time Requirements}

In terms of the requirements for real-time analysis in industrial inspection, it is necessary to consider the size of the product being inspected and the resolution of the camera(s) being used. \par
The authors of \cite{79_li13} developed custom smart cameras specifically for their use case. Each camera had a resolution of 1600x1200 pixels and a field of view of 800mm. This meant that six cameras were needed across the knitting machine for the final inline inspection system. 
In \cite{31_jun21}, the collected fabric images were cropped to 224x224 pixel patches by a sliding window moving horizontally by 32 steps and vertically by 24 steps. This resulted in each original image being discretised into 768 images for processing. The authors of \cite{56_ccelik14} note that the larger the image size, the more time is required for image processing. Similarly, this time is also proportional to the number of images extracted from one frame. \par
Carpet is typically manufactured in broadloom metres, widths of greater than 3.66m. Assuming the use of a camera that would result in 1 pixel per mm (at least), for an area of 3.66m by 1m, this would require more than 16 overlapping images to be processed before the next metre of carpet is rolled before the cameras. Thus processing speed and inference time of the detection model is important to keep up with the conveyored product. \par

\begin{table}[]
\scriptsize 
\caption{Average inference time of models (with ResNet18 backbones) for a single image with a Nvidia T4 GPU @ 5GHz and an Intel Xeon CPU @ 2.20GHz}
\label{table:speed}
\resizebox{\columnwidth}{!}{%
\begin{tabular}{|l||c|c|c|c|}
\hline
Model & \begin{tabular}[c]{@{}c@{}}No. Params\\ (Million)\end{tabular} & \begin{tabular}[c]{@{}c@{}}Size\\ (Mb)\end{tabular} & \begin{tabular}[c]{@{}c@{}}GPU\\ (s)\end{tabular} & \begin{tabular}[c]{@{}c@{}}CPU\\ (s)\end{tabular} \\ \hhline{-||----}
AE & 5.6 & 43 & 4.83 & 50.00 \\ \hhline{-||----}
FastFlow & 9.7 & 37 & 0.182 & 0.369 \\ \hhline{-||----}
PaDiM & 2.8 & 168 & 0.162 & 0.269 \\ \hhline{-||----}
PatchCore & 2.8 & 11 & \textbf{0.148} & 0.883 \\ \hhline{-||----}
Reverse Dist. & 18.2 & 69 & 0.157 & 0.275 \\ \hhline{-||----}
STFPM & 5.6 & 21 & 0.156 & \textbf{0.245} \\ \hhline{-||----}
\end{tabular}%
}
\end{table}

The results in Table \ref{table:speed} are from a standard GPU and CPU to reflect the typical computing units available in an industrial setting. The generative AE-based model yielded image results in the slowest time, most likely due to the CW-SSIM loss calculation. The feature embedding networks are also typically slower at performing inference than the student-teacher networks. This is with exception of PatchCore which, when running on a GPU, performs inference faster than all the other models. \par

\subsection{Areas for Improvement}

Observed shortcomings common across the evaluated models, including the best performing Reverse Distillation model, include poor detection of subtle, shade anomalies and poor detection of thin, linear anomalies. Addressing both of these points will be critical in achieving an effective inspection system for woollen tufted carpets. Both types of anomalies are very common, where blends of dyed wool fibre can be off-shade and the nature of yarn defects, when tufted into the final carpet, present themselves typically as thin linear anomalies. \par
Possible strategies for addressing these challenges include retaining the colour information of the images, either as input into the detection model or for pre-processing techniques. Another line of enquiry could be image distortions, such as stretching, that could help preserve anomalous features through a network's pooling layers. Finally, a component often added to popular object detection methods \cite{fast_rcnn, effnet} to improve performance are attention mechanisms. Attention mechanisms aim to generate new feature maps in between model layers to achieve selective focus and dampen weak features. \par

\section{Conclusions and Future Work}

In this paper, a custom carpet dataset was created in order to thoroughly evaluate state-of-the-art unsupervised anomaly detection methods for an automated inspection application. Textile classes within existing, publicly-available industrial datasets for anomaly detection, are typically of finely-woven, flat fabrics in which anomalies appear very differently than in tufted woollen carpets. \par
Grouped by their respective model types, on average, the student-teacher methods yielded the highest detection accuracies and lowest inference times compared to the generative and feature embedding models. It was also determined that training on a multi-class dataset yields comparable model performance and shows promise for industrial application. Finally, it was confirmed that input image patch size does have an effect on model performance and that, for this dataset, patches of 512 pixels yielded more accurate results than 256 pixels. \par
There are three major areas for further research that we have identified following this work. Firstly, selecting a model and adjusting it such that the model detection accuracy is as close as possible to 100\%. Currently, the best performing model, Reverse Distillation \cite{68_reverse_dist}, yields on average 92.2\% AUPRO and 64.8\% F1-Score across classes. Secondly, after selecting a suitable model, reducing the detection time of anomalies within large images, without necessarily relying on top-of-the-line computing equipment, to satisfy real-time requirements for an inline inspection system for broadloom carpets. Finally, investigating what pre and post image processing techniques could be used to improve detection performance, this includes histogram equalisation \cite{31_jun21, 79_li13}, adaptive thresholding techniques \cite{57_wei21, 79_li13}, and utilising morphology filters \cite{79_li13}. \par

\section{Acknowledgements}

This project was funded by Bremworth Carpets and Rugs, New Zealand.  Bremworth acknowledges the Co-Funding of this project by the Ministry for Primary Industries’ Sustainable Food and Fibre Futures Fund.

\printbibliography


\end{document}